\documentclass{article}

\usepackage{PRIMEarxiv}

\usepackage[utf8]{inputenc} 
\usepackage[T1]{fontenc}    
\usepackage{hyperref}       
\usepackage{url}            
\usepackage{booktabs}       
\usepackage{amsfonts}       
\usepackage{nicefrac}       
\usepackage{microtype}      
\usepackage{lipsum}
\usepackage{algorithm}
\usepackage{amsmath}
\usepackage{algorithmic}
\usepackage{multirow}
\usepackage{float}
\usepackage{amsfonts}
\usepackage{graphicx}
\usepackage{booktabs}
\usepackage{pifont}
\usepackage{fancyhdr}       
\usepackage{graphicx}       
\graphicspath{{media/}}     

\title{Structure-Aware Automatic Channel Pruning by Searching with Graph Embedding}

\author{
  Zifan Liu, Yuan Cao*, Yanwei Yu \\
  Ocean University of China \\
  Qingdao, China\\
  \texttt{\{lzf9793, cy8661, yuyanwei\}@ouc.edu.cn} \\
  \AND
  Heng Qi \\
  Dalian University of Technology \\
  Dalian, China \\
  \texttt{hengqi@dlut.edu.cn} \\
  \And
  Jie Gui \\
  Southeast University \\
  Nanjing, China\\
  \texttt{guijie@seu.edu.cn} \\
}

\begin{document}
\maketitle

\begin{abstract}
Channel pruning is a powerful technique to reduce the computational overhead of deep neural networks, enabling efficient deployment on resource-constrained devices. However, existing pruning methods often rely on local heuristics or weight-based criteria that fail to capture global structural dependencies within the network, leading to suboptimal pruning decisions and degraded model performance. To address these limitations, we propose a novel structure-aware automatic channel pruning (SACP) framework that utilizes graph convolutional networks (GCNs) to model the network topology and learn the global importance of each channel. By encoding structural relationships within the network, our approach implements topology-aware pruning and this pruning is fully automated, reducing the need for human intervention. We restrict the pruning rate combinations to a specific space, where the number of combinations can be dynamically adjusted, and use a search-based approach to determine the optimal pruning rate combinations. Extensive experiments on benchmark datasets (CIFAR-10, ImageNet) with various models (ResNet, VGG16) demonstrate that SACP outperforms state-of-the-art pruning methods on compression efficiency and competitive on accuracy retention.
\end{abstract}


\section{Introduction}
As deep learning continues to develop, the demand for efficient neural network deployments is at an all-time high. With the rise of edge computing and the proliferation of mobile and IoT devices, there is an urgent need to deploy lightweight models on resource-prioritized devices. However, Deep Neural Networks (DNNs), despite their outstanding performance \cite{Residual, kim2020spiking, lin2016scribblesup}, but it requires a lot of computational and storage overhead. And they also require significant memory resources, making them unsuitable for deployment in such environments.

To address this challenge, a lot of research has been conducted on compressed DNNs. Popular network compression methods include network pruning \cite{zhang2021structadmm, SRR-GR, xu2020layer, AutoCompress}, quantization \cite{HAWQ, EMQ, zhu2020towards, HAQ}, and knowledge distillation \cite{jin2021teachers, park2019relational, gou2021knowledge}. Among them, channel pruning has become a key technique. It can reduce the size and computational cost of DNNs without sacrificing accuracy \cite{he2017channel, luo2017thinet}. By removing less important channels (i.e., the output feature maps of the convolutional layer), channel pruning reduces the amount of computation during training and inference. This leads to more efficient models. However, traditional channel pruning techniques are usually manual or heuristic-based.  They require tuning hyperparameters or designing architecture-specific pruning rules. This manual intervention makes it difficult to scale channel pruning techniques to different network architectures or apply them to different tasks. And the lack of automation and adaptability limits the practical application of existing methods.

In addition, although existing pruning methods are effective \cite{ABCPruner, AMC, APRS}, they often fail to account for global structural dependencies between channels and layers. These methods ignore the overall performance capabilities of the model architecture. This can potentially lead to sub-optimal pruning decisions. This limitation is particularly evident in modern large-scale architectures. In these architectures, the network topology plays a crucial role in determining network performance.

To address the above problems, we propose a novel Structure-Aware and Automatic Channel Pruning (SACP) framework. In our approach, the overall architecture of a neural network is first represented as a graph structure. Each node in the graph corresponds to a convolutional, BN or pooling layer, and the edges represent the data flow relationships between layers.We then represent the channels corresponding to the convolutional layers as features of the nodes, and the graph is subsequently processed by a graphical convolutional network (GCNs). The GCNs captures the global topological structure of the model and encodes its architectural characteristics to generates a compact embedding. In parallel, we design a pruning search space which defines multiple combinations of hierarchical pruning rates. The size of this search space can be dynamically adjusted. It allows the size of the pruning space to be adapted to different tasks and model complexities. In order to guide the learning of channel importance in a meaningful way, we perform unsupervised comparison learning. The contrast learning occurs between the embedding of the original model and its pruned variants. It encourages the GCNs to generate structure-aware representations that reflect the performance potential of the architecture.

Based on the learned embeddings and importance scores, we employ a search-based strategy. This strategy automatically determines the optimal pruning configuration without relying on manual rules or adjustments. To the best of our knowledge, this is the first attempt to realize automatic pruning by searching from a network embedding space. Furthermore, we include a threshold constraint on the overall pruning rate of the model in the search space. This allows flexibility in controlling the overall pruning rate and striking a balance between model size and accuracy. In contrast to traditional approaches that rely only on local statistics, SACP utilizes global structural cues in a fully automated and data-driven manner. This enables our framework to seamlessly adapt to various network architectures and compression requirements. As a result, it is highly practical in real-world deployments.


The main contributions are summarized as follows. 
\begin{itemize}
    \item We propose a novel Structure-Aware Automatic Channel Pruning framework (SACP) that models the network topology as a graph and uses Graph Convolutional Networks (GCNs) to learn the performance of each model architecture, thus enabling topology-aware pruning without manual heuristics.
    \item  We design a dynamic search mechanism to optimize the combination of pruning rates in a constrained space, so that the pruning strategy can automatically adapt to different architectures and complexities, improving flexibility and efficiency.  
    \item Extensive experiments using multiple backbone models (ResNet, VGG16) on benchmark datasets (CIFAR-10 and ImageNet) show that SACP obtains superior balance on compression efficiency and accuracy retention compared with state-of-the-art pruning methods.
\end{itemize}

\section{Related Work}
Neural network compression has become crucial for deploying deep models on resource-constrained devices such as mobile phones, edge servers, and embedded systems. To reduce model size and computation, various techniques have been proposed, including efficient network architecture design \cite{Residual, Shufflenet, Mobilenets, Mobilenetv2}, weight quantization \cite{EMQ, Q-bert, PTQ, Zeroq}, knowledge distillation \cite{Distilling, sun2020contrastive, guo2023class}, low-rank decomposition\cite{denton2014exploiting, jaderberg2014speeding, sainath2013low, zhang2021structadmm}, and pruning\cite{l1-norm, AMC, luo2017thinet, Hrank}. Among them, channel pruning has attracted increasing attention due to its ability to remove redundancy while maintaining compatibility with existing hardware and inference frameworks. In the following, we focus on reviewing representative works on channel pruning.

\subsection{Channel Pruning}

Channel pruning is a widely used technique in neural network compression. It aims to reduce the number of channels (or feature maps) in a convolutional layer. This helps to speed up inference and reduce the memory footprint. Unlike unstructured pruning that removes individual weights, channel pruning produces a compact structured model. Unstructured pruning usually requires specialized hardware to achieve speedup. In contrast, channel pruning produces a model that is compatible with existing deep learning frameworks and devices.

\subsubsection{Traditional Channel Pruning Approaches}

Early works in channel pruning primarily rely on local importance metrics. These metrics evaluate the contribution of each channel based on simple statistics. One representative method is \textbf{L1-Norm Pruning} \cite{l1-norm}. It prunes channels with the smallest L1 norms of their convolutional filters, under the assumption that smaller weights contribute less to the output. \textbf{ThiNet} \cite{luo2017thinet} selects channels based on their influence on the next layer’s outputs. It formulates pruning as a feature reconstruction problem.

Another widely-used approach is \textbf{Network Slimming} \cite{liu2017learning}. It applies sparsity regularization to the scaling factors in Batch Normalization layers. Channels with small scaling factors are then removed after training. \textbf{HRank} \cite{Hrank} proposes evaluating channel importance by analyzing the rank of feature maps. It assumes that lower-rank features are less informative. Although these methods are simple and effective, they mainly focus on \textit{layer-wise or local statistics}. They fail to consider the \textit{global structural roles} of different channels within the entire network architecture.

\begin{figure*}[t]
	\centering
 \includegraphics[scale=0.5]{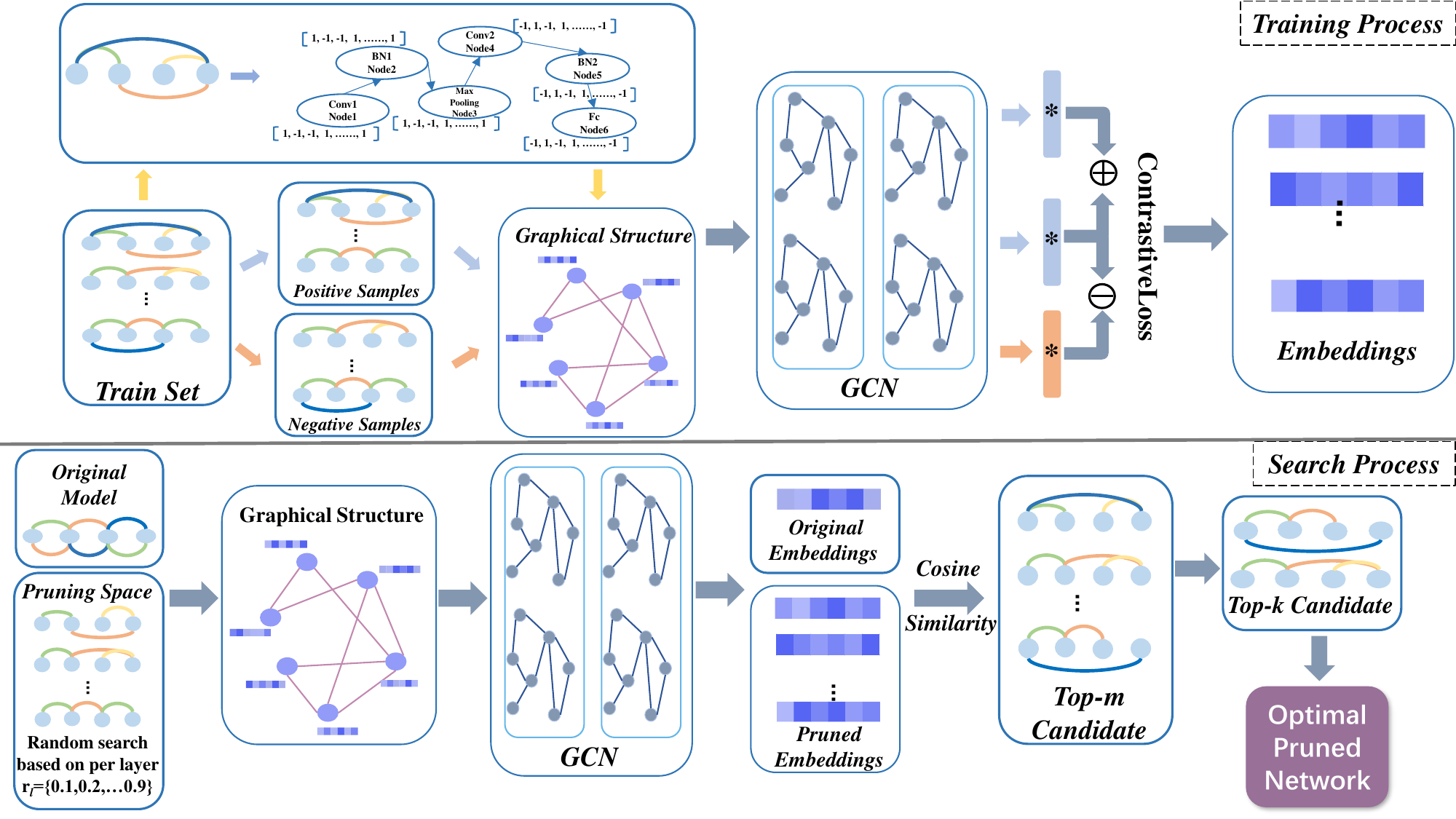}
 \caption{\normalsize The overall framework of the proposed SACP method. In the training phase (upper part), a set of pruned networks is generated using different pruning configurations to construct the training set. Positive samples are produced using L1-norm-based pruning, while negative samples are generated through random pruning. Each configuration is converted into a graphical structure and input into a GCNs encoder to learn structure-aware embeddings. A contrastive loss is applied to optimize the GCNs. In the search phase (bottom part), the original model and a large pool of pruning candidates are encoded into the embedding space using the trained GCNs. Cosine similarity is computed between each pruned candidate and the original model embedding. The top-$m$ candidates with the highest similarity are selected and lightly fine-tuned. Finally, the top-$k$ candidates are fully retrained, and the best-performing one is selected as the final pruned network.}
	\label{framework}
\end{figure*}

\subsubsection{Automatic Pruning Approaches}

To overcome the limitations of manually crafted heuristics, several learning-based and automated pruning methods have been proposed. \textbf{AMC}~\cite{AMC} formulates the pruning process as a reinforcement learning problem, where a pruning agent sequentially determines the optimal compression policy for each layer. \textbf{MetaPruning}~\cite{Metapruning} utilizes a meta-learning approach to learn a pruning strategy generator that can generalize across network architectures. \textbf{ABCPruner}~\cite{ABCPruner} proposes an artificial bee colony algorithm-based method to automatically search for optimal channel pruning structures by reducing the combination space and enabling end-to-end fine-tuning. \textbf{AGMC}~\cite{AGMC}  proposes an automatic model compression method by learning DNN structural information with graph neural networks and optimizing pruning policies via reinforcement learning. \textbf{APIB}~\cite{APIB} proposes an automatic network pruning method based on the Information Bottleneck (IB) principle, utilizing HSIC Lasso to optimize channel selection and pruning ratios while reducing reliance on heuristics.

\newcommand{\cmark}{\ding{51}} 
\newcommand{\xmark}{\ding{55}} 
\newcommand{\pmark}{\textbf{--}} 

\begin{table}[t]
\centering
\caption{Comparison of representative pruning methods.}
\label{tab:pruning_comparison}
\small
\setlength{\tabcolsep}{3.8pt}
\begin{tabular}{lcccccc}
\toprule
\textbf{Method} & \textbf{Struct.} & \textbf{Auto} & \textbf{Search} & \textbf{CL} & \textbf{Adapt.} \\
\midrule
L1-Norm \cite{l1-norm}              & \xmark & \xmark & \xmark & \xmark & \xmark \\
ThiNet \cite{luo2017thinet}        & \xmark & \xmark & \xmark & \xmark & \xmark \\
Network Slimming \cite{liu2017learning}     & \xmark & \xmark & \xmark & \xmark & \xmark \\
HRank \cite{Hrank} & \xmark & \xmark & \xmark & \xmark & \xmark \\
AMC \cite{AMC} & \xmark & \cmark & \cmark & \xmark & \pmark \\
MetaPruning\cite{Metapruning} & \xmark & \cmark & \cmark & \xmark & \pmark \\
ABCPruner \cite{ABCPruner} & \xmark & \cmark & \cmark & \xmark & \cmark \\
AGMC \cite{AGMC} & \cmark & \cmark & \cmark & \xmark & \pmark \\
\textbf{SACP (Ours)} & \cmark & \cmark & \cmark & \cmark & \cmark \\
\bottomrule
\end{tabular}
\vspace{0.5em}
\caption*{\footnotesize \textbf{Struct.}: Structural information; \textbf{Auto}: Automated pruning; \textbf{Search}: Search-based strategy; \textbf{CL}: Contrastive learning; \textbf{Adapt.}: Adaptability to different architectures.}
\end{table}

As shown in Table \ref{tab:pruning_comparison}, many representative methods rely on local importance metrics or heuristic rules without incorporating structural context (e.g., L1-Norm, ThiNet, Network Slimming). Even search-based automatic methods, such as MetaPruning and ABCPruner, do not fully utilize topological information or comparative representation learning. In contrast, our proposed SACP framework integrates structural modeling, automated search, and comparative learning in a unified pipeline and shows strong adaptability across various network architectures.

\subsection{Graph Convolutional Networks}
Graph Convolutional Networks (GCNs) \cite{GCN} have been widely used in various tasks. This is due to their ability to capture structural relationships in graph data \cite{kim2020spiking, velivckovic2017graph}. In particular, GCNs are used in Neural Architecture Search (NAS) \cite{you2020graph, cai2018proxylessnas}. They are also applied in model performance prediction \cite{liu2018progressive} and architectural representation learning \cite{he2021automl}. In these applications, the network is viewed as a graph. The graph is composed of interconnected components, such as layers and blocks. However, their application in channel pruning remains underexplored. Unlike previous approaches, our approach utilizes GCNs which helps explicitly model the network topology and learn channel importance from a global, structure-aware perspective.

\subsection{Contrastive Learning for Representation Optimization}

Contrastive learning has recently gained traction as a powerful unsupervised representation learning method. Techniques like SimCLR \cite{chen2020simple} and MoCo \cite{he2020momentum} have demonstrated its effectiveness in learning discriminative features by maximizing agreement between positive pairs while minimizing the similarity between negative pairs. This approach allows models to learn useful feature representations without relying on labeled data. While contrastive learning is commonly applied to image-level tasks, its application has been extended to other domains, including graph embeddings \cite{hamilton2017inductive} and architecture representation learning \cite{Metapruning}. In our work, we integrate contrastive learning to align the embedding of the original model and its pruned variants. This process encourages the pruning framework to preserve essential structural information while optimizing the model, ultimately leading to more efficient pruning that maintains performance.

\section{Methodology}

In this section, we introduce the proposed \textbf{Structure-Aware Automatic Channel Pruning (SACP)} framework in detail. The general framework of the proposed SACP method is shown in Figure \ref{framework}. SACP aims to efficiently prune convolutional neural networks (CNNs) by modeling the architectural structure explicitly and guiding the pruning process through a GCNs-based encoder trained via unsupervised contrastive learning. The goal is to automatically determine effective layer-wise pruning configurations that reduce computation while preserving performance.

\subsection{Problem Definition and Notations}

Given a CNN $\mathcal{F}$ composed of $L$ layers, where each layer $l$ has $C_l$ output channels, our objective is to find a set of layer-wise pruning ratios $\mathbf{r} = [r_1, r_2, ..., r_L]$ with $r_l \in [0,1]$, such that the pruned model $\mathcal{F}'$ satisfies a desired global pruning ratio while maintaining competitive accuracy. The global pruning ratio is defined as

\begin{equation}
R(\mathbf{r}) = \frac{\sum_{l=1}^{L} r_l \cdot C_l}{\sum_{l=1}^{L} C_l},
\end{equation}
and we enforce the constraint $R(\mathbf{r}) \geq \tau$, where $\tau$ is a user-defined threshold. The challenge lies in identifying $\mathbf{r}$ in a high-dimensional, discrete search space that balances compression and performance, without manual tuning or task-specific heuristics.

\subsection{Graph-Based Network Representation}

To enable structure-aware pruning, we represent a neural network architecture as a Directed Acyclic Graph (DAG) $G = (V, E)$, where each node $v_i \in V$ denotes a specific computational layer, and each directed edge $e_{ij} \in E$ encodes the information flow from layer $i$ to layer $j$. This formulation captures both sequential layer relationships and complex topological dependencies, such as skip connections, branching, and merging operations, which are common in modern architectures (e.g., ResNet). For standard feedforward layers, we connect each layer to its immediate successor. In the case of non-sequential modules such as residual blocks, we introduce edges that reflect long-range dependencies to preserve functional pathways in the graph. As a result, the graph topology explicitly models both shallow and deep hierarchical information in the original network.

To reflect the pruning status of each layer, we associate each node $v_i$ with a binary feature vector $x_i \in \{0,1\}^d$, where $d$ is the maximum number of output channels among all layers:
\[
d = \max_{l \in \{1, 2, \dots, L\}} C_l.
\]
The feature vector $x_i$ represents the channel-wise mask for layer $i$: if the $j$-th channel is retained after pruning, then $x_i[j] = 1$; otherwise, $x_i[j] = -1$. Formally, for a given pruning vector $\mathbf{r}$ and corresponding preserved channel set $\mathcal{P}_i$ for layer $i$:
\[
x_i[j] =
\begin{cases}
1, \& \text{if } j \in \mathcal{P}_i, \\
-1, \& \text{otherwise}.
\end{cases}
\]

To handle layers with fewer than $d$ channels, we zero-pad the feature vector to maintain consistent input dimensions across nodes. For Batch Normalization layers, we assign them the same feature vector as their preceding convolutional layer to preserve semantic coherence. Fully Connected (FC) layers are not subject to pruning in our setting and are therefore assigned an all-one vector: $x_i = \mathbf{1}_d$.

This graph formulation encodes both the architectural topology and the fine-grained channel pruning state of the network. Such a representation enables the GCNs to jointly learn from the hierarchical structure and the pruning configuration, making it possible to reason about the global impact of pruning decisions at the architecture level.

\subsection{Structured Pruning Dataset Generation}

To effectively train the GCNs encoder for learning structure-aware pruning policies, we construct two sets of pruned network configurations: one for training and one for search. Each configuration corresponds to a specific pruning strategy applied across all prunable layers, resulting in a unique network architecture and graph-based structural representation.

Let the original network consist of $L$ prunable layers, each with $C_l$ output channels for $l = 1, 2, \dots, L$. We define a pruning vector $\mathbf{r} = [r_1, r_2, \dots, r_L]$, where $r_l \in [0, 1)$ denotes the proportion of channels to be pruned in layer $l$. The number of remaining channels in layer $l$ is given by:
\begin{equation}
C_l^{\text{pruned}} = \left\lfloor (1 - r_l) \cdot C_l \right\rfloor.
\end{equation}
To reduce the search space while maintaining sufficient granularity, we discretize each $r_l$ into a fixed set:
\begin{equation}
r_l \in \mathcal{R} = \{0, 0.1, 0.2, \dots, 0.9\}.
\end{equation}
This results in a pruning configuration space of size $|\mathcal{R}|^L$ for an $L$-layer network.

During GCNs training, our goal is to improve the encoder's ability to learn structure-aware representations. To this end, we randomly draw $N$ pruning vectors $\{\mathbf{r}^{(1)}, \dots, \mathbf{r}^{(N)}\}$ from the entire search space without imposing any global pruning ratio constraints. This ensures that the structural diversity in the training samples is maximized, allowing the GCNs to learn robust, generalizable embeddings that reflect structural differences and pruning patterns.

In contrast, in the search phase, we construct a pruned spatial search set that targets practical deployment. Unlike the training set, which emphasizes diversity, the search set is generated under explicit constraints to ensure a practical level of compression. Specifically, we define the global pruning ratio $R(\mathbf{r})$ as the ratio of pruned channels relative to the total number of original channels, and enforce a threshold $\tau$ to ensure a minimum level of compression:
\begin{equation}
R(\mathbf{r}) = \frac{\sum_{l=1}^{L} r_l \cdot C_l}{\sum_{l=1}^{L} C_l} \geq \tau.
\end{equation}
This limit effectively eliminates configurations that are too close to the unpruned model, which otherwise provide negligible efficiency gains. By enforcing a compression lower bound, we can direct the search process to explore meaningful design points that strike a balance between reducing parameters and maintaining performance. This also ensures that the pruned candidates evaluated during the inference process are feasible and competitive in terms of efficiency in real-world deployments.

Each pruning configuration, whether used for training or search, is transformed into a graph $G^{(i)}$ with binary node features, following the protocol in Section~3.2. The training configurations form the foundation for contrastive learning of the GCNs encoder, while the constrained candidate set is used during inference to guide the selection of optimal pruning strategies.

\subsection{GCNs-Based Contrastive Embedding Learning}

After constructing graph representations for pruned networks, we train a graph encoder $\mathcal{G}_\theta$ based on Graph Convolutional Networks (GCNs) to map each network graph into a low-dimensional embedding. These embeddings are designed to capture both the architectural topology and the pruning configuration, enabling the encoder to reason about structural differences between networks. The ultimate goal is for the embedding space to reflect the relative quality of different pruning strategies in a structure-aware manner.

Since explicit performance labels are not available at the early pruning stage, we use a supervised contrast learning framework to train the encoder. In each training iteration, we construct an anchor-positive-negative triad, which consists of an anchor graph sampled from the training set, a positive graph obtained through pruning based on the L1 criterion, and a negative graph generated through random pruning. Positive samples are expected to preserve high-frequency channels and maintain reasonable performance, while negative samples ignore structural signals and tend to break important connections. All graphs are converted into binary feature DAGs (as described in Section 3.2) and the corresponding embeddings are obtained via the GCNs encoder.

Given a batch of $N$ anchor-contrast pairs, we first compute the pairwise scaled cosine similarity between anchor embeddings $z_1$ and contrast embeddings $z_2$:
\begin{equation}
\text{sim}(z_1, z_2) = \frac{z_1 \cdot z_2^\top}{\tau},
\end{equation}
where $\tau$ is a temperature hyperparameter that controls the sharpness of the similarity distribution. To improve numerical stability, we subtract the row-wise maximum similarity before applying exponentials.

We then define a contrastive loss for each pair based on softmax-normalized similarities. Let $y_i \in \{0,1\}$ indicate whether the $i$-th anchor-contrast pair is positive or negative. The loss for the $i$-th pair is
\begin{equation}
\mathcal{L}_{\text{CL}}^{(i)} = -\log \left( \frac{\exp(\text{sim}(z_1^{(i)}, z_2^{(i)}) - \max_j \text{sim}(z_1^{(i)}, z_2^{(j)}))}{\sum_{j=1}^{N} \exp(\text{sim}(z_1^{(i)}, z_2^{(j)}) - \max_j \text{sim}(z_1^{(i)}, z_2^{(j)})) + \epsilon} \right),
\end{equation}
where $\epsilon$ is a small constant added for numerical safety. This formulation encourages the encoder to assign higher similarity to semantically aligned structures and lower similarity to dissimilar ones.

To form the final training objective, we compute the contrastive loss separately for positive and negative anchor-contrast pairs. Let $P$ and $N$ denote the index sets of positive and negative samples within a batch, respectively. For each anchor-positive pair $(z_1^{(i)}, z_2^{(i)})$ with $i \in P$, and each anchor-negative pair $(z_1^{(j)}, z_2^{(j)})$ with $j \in N$, we compute the contrastive loss $\mathcal{L}_{\text{CL}}^{(i)}$ using the softmax-normalized similarity-based formulation defined above. The average losses over positive and negative sets are defined as

\begin{equation}
\mathcal{L}_{\text{pos}} = \frac{1}{|P|} \sum_{i \in P} \mathcal{L}_{\text{CL}}^{(i)}, \quad
\mathcal{L}_{\text{neg}} = \frac{1}{|N|} \sum_{j \in N} \mathcal{L}_{\text{CL}}^{(j)}.
\end{equation}
The final training objective is then the sum of the two components:
\begin{equation}
\mathcal{L}_{\text{total}} = \mathcal{L}_{\text{pos}} + \mathcal{L}_{\text{neg}}.
\end{equation}

Through this contrastive training strategy, the GCNs encoder learns to distinguish pruning configurations that preserve meaningful structural characteristics from those that degrade performance. This enables the model to embed architectures in a way that reflects their structural importance and expected pruning quality, without requiring full network retraining or inference during the training phase.

\begin{algorithm}[t]
\caption{Structure-Aware Automatic Channel Pruning (SACP)}
\label{alg:SACP}
\begin{algorithmic}[1]
\REQUIRE Original network $\mathcal{F}$, pruning threshold $\tau$, pruning granularity $\Delta$
\STATE Construct graph $G_{\mathcal{F}}$ with binary pruning features
\STATE Generate pruning configurations $\{\mathbf{r}^{(i)}\}$ with $R(\mathbf{r}^{(i)}) \geq \tau$
\FOR{each configuration $\mathbf{r}^{(i)}$}
    \STATE Construct graph $G^{(i)}$ and positive/negative variants
    \STATE Encode embeddings $z$, $z^+$, $z^-$ with GCNs
    \STATE Update GCNs via contrastive loss $\mathcal{L}_{\text{CL}}$
\ENDFOR
\STATE Encode original model embedding $z_\text{orig}$
\STATE Sample large candidate set $\{\mathbf{r}^{(j)}\}$ satisfying threshold
\FOR{each candidate}
    \STATE Encode $z^{(j)} = \mathcal{G}_\theta(G^{(j)})$ and compute similarity $s_j$
\ENDFOR
\STATE Select top-$m$ candidates with highest $s_j$
\STATE Lightly fine-tune top-$m$ and evaluate performance
\STATE Select top-$k$ candidates for full training
\RETURN Final pruned model with best accuracy
\end{algorithmic}
\end{algorithm}

\subsection{Pruning Strategy Selection and Final Optimization}

Once the GCNs encoder $\mathcal{G}_\theta$ is trained, it can be used to evaluate the pruning quality of new configurations directly, without the need for retraining the model. To achieve this, we first generate a large pool of candidate pruning vectors $\{\mathbf{r}^{(j)}\}$, ensuring that each configuration satisfies the global pruning constraint $R(\mathbf{r}^{(j)}) \geq \tau$, where $\tau$ is a user-defined threshold. The total pruning ratio $R(\mathbf{r})$ is defined as

\begin{equation}
R(\mathbf{r}) = \frac{\sum_{l=1}^{L} r_l \cdot C_l}{\sum_{l=1}^{L} C_l},
\end{equation}
where $C_l$ is the number of channels in layer $l$, and $r_l$ is the pruning ratio for layer $l$. This ensures that the pruning configurations under consideration achieve a sufficient level of compression while maintaining meaningful model structure.

For each pruning candidate $\mathbf{r}^{(j)}$, we convert it into a binary feature graph $G^{(j)}$ following the same process described in Section 3.2. The graph $G^{(j)}$ is then passed through the trained GCNs encoder, which produces an embedding vector $z^{(j)}$ that captures the structural representation of the pruned network. To assess the quality of the pruning configuration, we compute the cosine similarity between the embedding of the candidate $z^{(j)}$ and the embedding of the original unpruned model $z_{\text{orig}}$:

\begin{equation}
s^{(j)} = \cos(z^{(j)}, z_{\text{orig}}) = \frac{z^{(j)} \cdot z_{\text{orig}}}{\|z^{(j)}\|_2 \cdot \|z_{\text{orig}}\|_2}.
\end{equation}

The similarity score $s^{(j)}$ serves as an indicator of how closely the pruning configuration resembles the original model in terms of structural preservation. A higher similarity score suggests that the pruned model retains more of the original network’s important features, which is expected to result in better performance after pruning.

We select the top-$m$ configurations from the candidate models that have the highest similarity to the original model. These candidate configurations are lightly fine-tuned to further evaluate their performance. After fine-tuning, we select the top-$k$ configurations based on their validation accuracy and fully train them until convergence. The final pruned model is chosen based on the test performance of these top-$k$ configurations. The entire process of pruning strategy selection and optimization is summarized in Algorithm~\ref{alg:SACP}.

\begin{table*}[ht]
\centering
\renewcommand{\arraystretch}{1.2}
\setlength{\tabcolsep}{3pt}
\caption{Pruning results of VGG-16, ResNet-18, and ResNet-56 on CIFAR-10.}
\begin{tabular}{l l c c c c c}
\hline
\textbf{Model} & \textbf{Method} & \textbf{Top-1 Accuracy} & \textbf{FLOPs} & \textbf{Pruned FLOPs} & \textbf{Parameters} & \textbf{Pruned Parameters} \\
\hline

\multirow{5}{*}{VGG-16} 
& Base & 92.60\% & 314.31M & 0.00\% & 14.73M & 0.00\% \\
& GAL-0.05 & 90.86\% & 189.49M & 39.71\% & 3.36M & 77.19\% \\
& Hrank & 91.26\% & 108.61M & 65.44\% & 2.64M & 82.08\% \\
& ABCPruner-80\% & 93.08\% & 82.81M & 73.65\% & 1.67M & 88.68\% \\
& \textbf{SACP (ours)} & 92.76\% & 55.86M & 82.23\% & 1.16M & 92.12\% \\

\hline
\multirow{3}{*}{ResNet-18} 
& Base & 92.32\% & 557.89M & 0.00\% & 11.17M & 0.00\% \\
& OTOv2 & 92.86\% & 113.25M & 79.70\% & - & - \\
& ATO & 94.51\% & 112.69M & 79.80\% & - & - \\
& \textbf{SACP (ours)} & 93.41\% & 83.76M & 84.99\% & 0.82M & 92.67\% \\

\hline
\multirow{7}{*}{ResNet-56} 
& Base & 94.58\% & 127.93M & 0.00\% & 0.86M & 0.00\% \\
& GAL-0.6 & 92.98\% & 78.30 & 38.79\% & 0.75M & 12.79\% \\
& FPGM & 92.93\% & 60.64M & 52.60\% & 0.47M & 45.35\% \\
& HRank & 92.17\% & 63.96M & 50.00\% & 0.49M & 43.02\% \\
& ABCPruner-70\% & 93.23\% & 58.54M & 54.24\% & 0.39M & 54.65\% \\
& ATO & 93.74\% & 57.57M & 55.00\% & - & - \\
& \textbf{SACP (ours)} & 90.06\% & 28.07M & 78.06\% & 0.25M & 70.48\% \\

\hline
\end{tabular}
\label{cifar10_results}
\end{table*}

\section{Experiments}

In this section, we evaluate the performance of the proposed SACP method on standard image classification tasks. We first describe the datasets, experimental setup, and baselines used for comparison. Then, we present the results of our experiments, including the comparison with state-of-the-art pruning methods and ablation studies.

\subsection{Datasets}

We conduct experiments on two widely used image classification datasets: CIFAR-10 and ILSVRC 2012. For both datasets, we use standard data augmentation techniques, including random cropping and horizontal flipping, during training.

\textbf{CIFAR-10:} This dataset consists of 60,000 32x32 color images in 10 classes, with 50,000 training images and 10,000 test images. The dataset is commonly used for evaluating image classification algorithms and is divided into 5,000 training and 1,000 test images per class.

\textbf{ImageNet (ILSVRC 2012):} The dataset contains over 1.2 million images across 1000 training categories and 50,000 validation images. These images vary in size, but are typically resized to 224x224 pixels before training. Due to the diversity and complexity of the dataset, the ILSVRC 2012 version of ImageNet is a challenging benchmark for image classification tasks.


\subsection{Experimental Setup}

For training the SACP method, we conduct experiments on two standard image classification datasets: CIFAR-10 and ILSVRC 2012. The models are implemented using PyTorch and trained on a single NVIDIA GeForce RTX 3090 GPU.

\textbf{Training Settings:} For CIFAR-10, the learning rate is set to 0.01, the batch size is 128, and the models are trained for 200  to 250 epochs. The optimizer used is stochastic gradient descent (SGD) with momentum 0.9 and weight decay of $5 \times 10^{-4}$ to prevent overfitting. The learning rate follows a cosine annealing schedule, gradually decaying throughout the training process. Fine-tuning is performed after pruning to recover the model’s performance. For ImageNet, the learning rate is set to 0.1, and the batch size remains 512. The model is trained for 90 epochs, using the same weight decay of $5 \times 10^{-4}$ and cosine annealing for learning rate adjustment. As with CIFAR-10, pruning is followed by fine-tuning to restore performance.

\textbf{GCNs Training and Selection Strategy.}  
For each model, we construct a training dataset of pruned networks with varying layer-wise pruning rates. Each pruning configuration is converted into a graph representation with binary node features. The GCNs encoder is trained using contrastive learning for 10 to 20 epochs with a fixed batch size of 32. After training, a large candidate pool of pruning configurations is evaluated by computing the cosine similarity between each candidate’s embedding and that of the original model. The top-$m$ candidates are selected for lightweight fine-tuning, and the top-$k$ configurations, chosen based on validation accuracy, are fully retrained to convergence. Final performance is reported based on the best test accuracy. The specific settings of top-$m$ and top-$k$ for each model will be detailed in the subsequent results section to facilitate reproducibility and clarity.

\textbf{Evaluation Metrics:} The performance of the models is evaluated using three key metrics: accuracy, the FLOPs pruned rate, and the parameter pruned rate. Accuracy is the primary metric, calculated as the percentage of correct predictions on the test set. The FLOPs pruned rate measures the reduction in computational complexity after pruning, calculated as the percentage of floating-point operations (FLOPs) reduced compared to the original model. Similarly, the parameter prune rate evaluates the compression of the model by calculating the percentage of pruned parameters relative to the total number of parameters in the original model. These metrics allow us to assess the effectiveness of pruning in terms of model performance, computational cost reduction, and parameter compression.

\subsection{Results on CIFAR-10}

For the CIFAR-10 dataset, we perform experiments on three popular architectures: VGG-16, ResNet-18, and ResNet-56. We apply the SACP method on these networks and evaluate the pruning results, focusing on the top-1 accuracy before and after pruning. The results demonstrate that SACP achieves significant pruning while retaining competitive accuracy.

\textbf{VGG-16.} For VGG-16 on CIFAR-10, we train the GCNs using 10,000 pruning configurations and evaluate 1,000,000 candidates during inference under a global pruning ratio constraint of over 60\%. We select the top-150 candidates based on cosine similarity and fully retrain the top-10 based on validation accuracy. SACP removes 82.23\% of FLOPs, and 92.12\% of parameters, while slightly improving the classification accuracy to 92.76\% compared to the baseline. This demonstrates that SACP can significantly compress VGG-16 without performance degradation, making it more suitable for deployment on resource-constrained devices.

\textbf{ResNet-18.} For ResNet-18, GCNs was trained on 10,000 pruning configurations and evaluated on 1 million candidates during inference to ensure a global pruning rate higher than 70\%. We selected the top 150 candidates based on cosine similarity and fully retrained the top 10 candidates using validation accuracies. SACP reduces FLOPs by approximately 85\% and compresses model parameter sizes by over 92\% while maintaining 93.41\% top-1 accuracy over the pruned baseline. These results show that SACP is able to aggressively prune the residual network without compromising accuracy.

\textbf{ResNet-56.} For ResNet-56, we used 50,000 pruning configurations to train the GCNs and evaluated 5 million candidates during inference with a global pruning rate constraint of over 30\%. The top 300 candidates were filtered based on similarity, and the top 10 candidates were fully retrained. SACP reduced FLOPs by over 78\%, and compressed about 70\% of the model parameters while causing only a small accuracy degradation.These results highlight SACP's robustness to deeper architectures, as well as the ability to maintain performance.

\subsection{Ablation Study}
To investigate the contribution of each component in the SACP framework, we performed controlled ablation experiments on two key variants using ResNet-18 on CIFAR-10. SACP-1 denotes replacing the unsupervised contrastive learning in GCNs training with a supervised regression objective, where a linear layer predicts the validation accuracy of each pruning configuration. SACP-2 removes the top-$k$ refinement stage and directly selects the best-performing model from the top-$m$ pool. 

\begin{table}[h]
\centering
\caption{Ablation results of SACP on CIFAR-10 with ResNet-18.}
\label{tab:ablation}
\tabcolsep=0.17cm
\begin{tabular}{l c c c}
\toprule
\textbf{Method} & \textbf{Top-1 Acc. (\%)} & \textbf{FLOPs Red. (\%)} & \textbf{Para Red. (\%)}\\
\midrule
SACP-1      & 88.73 & \textbf{91.49} & 91.32 \\
SACP-2     & 88.91 & 80.46 & 92.47\\
\textbf{SACP} & \textbf{93.41} & 84.99 & \textbf{92.67} \\
\bottomrule
\end{tabular}
\label{tab:ablation}
\end{table}
As shown in Table~\ref{tab:ablation}, both variants result in a performance drop compared to the full SACP model. SACP-1 achieves the highest FLOPs reduction (91.49\%) but suffers from the lowest Top-1 accuracy (88.73\%), indicating that the supervised prediction fails to generalize pruning quality effectively. SACP-2 improves accuracy slightly but still underperforms. In contrast, the full SACP achieves the best Top-1 accuracy (91.54\%) while maintaining a strong balance in FLOPs (84.99\%) and parameter reduction (92.67\%), demonstrating the effectiveness of contrastive embedding and the two-stage selection strategy.

\subsection{Hyperparameter Analysis}

\begin{figure}[t]
    \centering
    \mbox{
    \hspace{-0.18in}
    \includegraphics[width=1.75in]{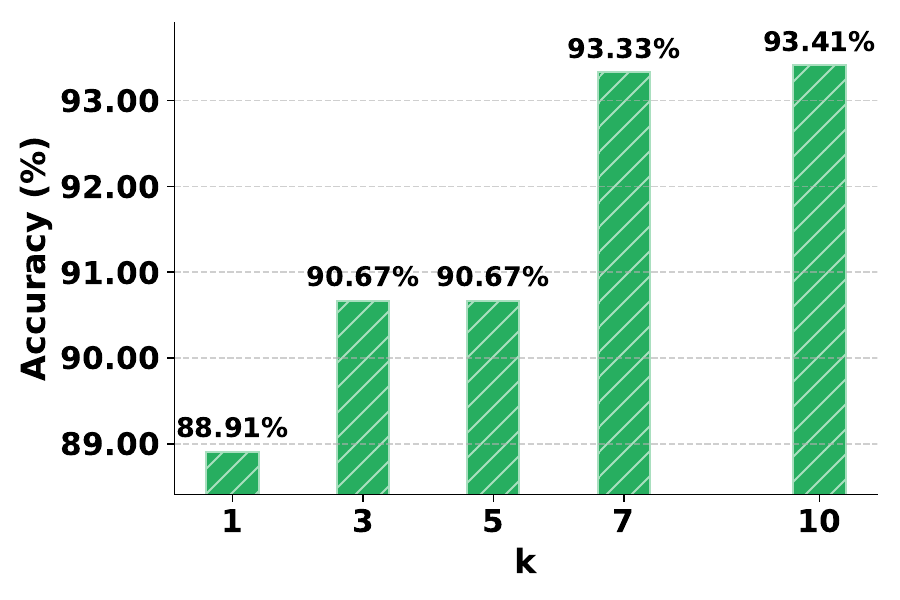}
\includegraphics[width=1.75in]{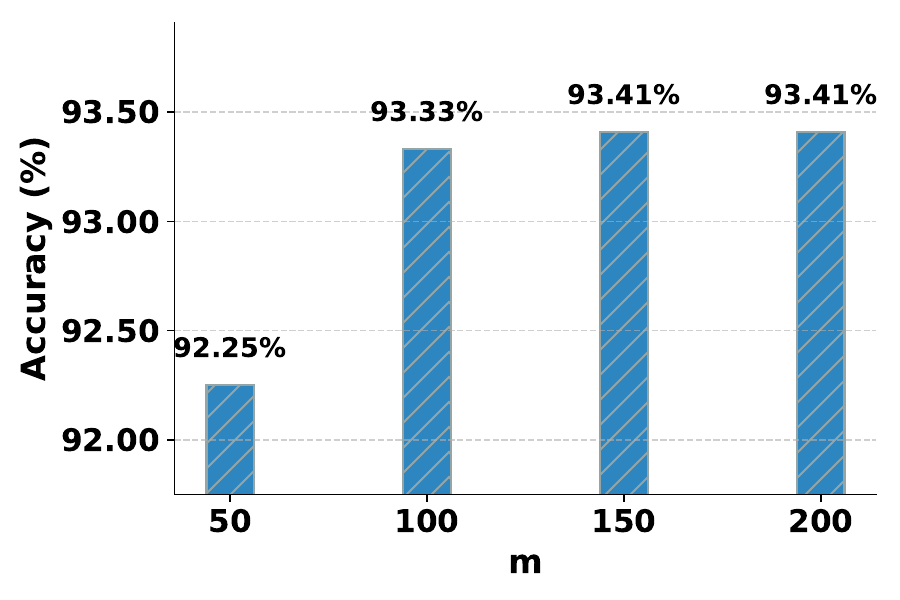}
    }
    \caption{Effect of $k$ and $m$ values on pruning performance with ResNet-18 on CIFAR-10.}
    \label{fig:topk}
\end{figure}

We analyze two key hyperparameters in SACP: the number of candidates retained after similarity-based ranking ($m$) and the number of candidates retrained for final selection ($k$). These two parameters directly affect the trade-off between computational overhead and pruning effectiveness. To evaluate the impact of $k$, we fix the variation of $m=150$ and $k$ to range from 1 to 10. As shown in Figure \ref{fig:topk}, increasing $k$ consistently improves the final accuracy. Larger $k$ allows the framework to avoid suboptimal selection by exploring more candidates during retraining. This suggests that a broader refinement phase is beneficial, especially in cases where similarity-based rankings may not perfectly reflect true performance. Similarly, to study the impact of $m$, we fix $k=10$ and gradually increase $m$. Expanding the pool of candidates improves the diversity and quality of the final choice set, leading to more accurate pruning decisions. However, we observe that the performance gain saturates beyond $m=150$, suggesting that the GCN encoder already provides highly discriminative embeddings for effective pruning with moderate candidate pool size. These results validate the importance of $k$ and $m$ and empirically support the choice of $k=10$ and $m=150$ as an effective balance between pruning accuracy, ranking stability, and computational cost.

\section{Conclusion}
In this paper, we propose a novel Structure-Aware Automatic Channel Pruning method called SACP. By modeling the network architecture as a graph, we utilize Graph Convolutional Networks (GCNs) to learn structure-aware embeddings for efficient pruning. We use unsupervised contrast learning to optimize GCNs to infer pruning policies directly from the network structure. Experimental results on CIFAR-10 and ImageNet show that SACP outperforms state-of-the-art pruning methods in terms of compression efficiency and is competitive in terms of accuracy preservation. Future work focuses on extending SACP to larger models and exploring more advanced pruning strategies.


\bibliographystyle{unsrt}
\bibliography{sample-base}

\end{document}